\documentclass[sigconf,screen]{acmart}

\usepackage{multirow}
\usepackage{booktabs}
\usepackage{caption}

\setcopyright{acmlicensed}
\copyrightyear{2025}
\acmYear{2025}
\acmDOI{XXXXXXX.XXXXXXX}
\acmConference[MM '25]{Proceedings of the 33rd ACM International Conference on Multimedia}{October 27--31, 2025}{Dublin, Ireland}

\NewDocumentCommand{\zzx}{ mO{} }{\textcolor{yellow}{\textsuperscript{\textit{zzx}}\textsf{\textbf{\small[#1]}}}}

\setcopyright{none}
\acmConference[]{}{}{}
\acmBooktitle{}
\renewcommand\footnotetextcopyrightpermission[1]{}

\renewcommand{\shortauthors}{Wang et al.}

\begin{document}

\pagestyle{plain}

%% add 

\author{Zihan Wang, Lu Yuan, Zhengxuan Zhang, Qing Zhao}
\affiliation{%
  \institution{Communication University of China}
  \city{Beijing}
  \country{China}
}

\title{Bridging Cognition and Emotion: Empathy-Driven Multimodal Misinformation Detection}

\begin{abstract}
    In the digital era, social media has become a major conduit for information dissemination, yet it also facilitates the rapid spread of misinformation. Traditional misinformation detection methods primarily focus on surface-level features, overlooking the crucial roles of human empathy in the propagation process. To address this gap, we propose the Dual-Aspect Empathy Framework (DAE), which integrates cognitive and emotional empathy to analyze misinformation from both the creator and reader perspectives. By examining creators’ cognitive strategies and emotional appeals, as well as simulating readers’ cognitive judgments and emotional responses using Large Language Models (LLMs), DAE offers a more comprehensive and human-centric approach to misinformation detection. Moreover, we further introduce an empathy-aware filtering mechanism to enhance response authenticity and diversity. Experimental results on benchmark datasets demonstrate that DAE outperforms existing methods, providing a novel paradigm for multimodal misinformation detection.
\end{abstract}

\maketitle

%!TEX root = ../main.tex
\section{Introduction}
\label{sec:introduction}

In the wave of the digital era, social media platforms have become the primary arena for information exchange~\cite{chen2023spread,mai2023dynamic}, yet they have also provided fertile ground for the large-scale spread of misinformation~\cite{abdali2024multi,zhou2020survey, shu2017fake,yindetecting}. Misinformation, like a digital plague, propagates through cyberspace at unprecedented speed and scale, not only disrupting public perception of truth but also profoundly affecting the formation of social consensus and the development of public policies~\cite{guo2019future,su2020motivations}. Faced with this complex and urgent challenge, establishing efficient and accurate misinformation identification mechanisms has become a key task in maintaining the health of the information ecosystem and social stability~\cite{mridha2021comprehensive}.

\begin{figure}
    \centering
    \includegraphics[width=0.8\linewidth]{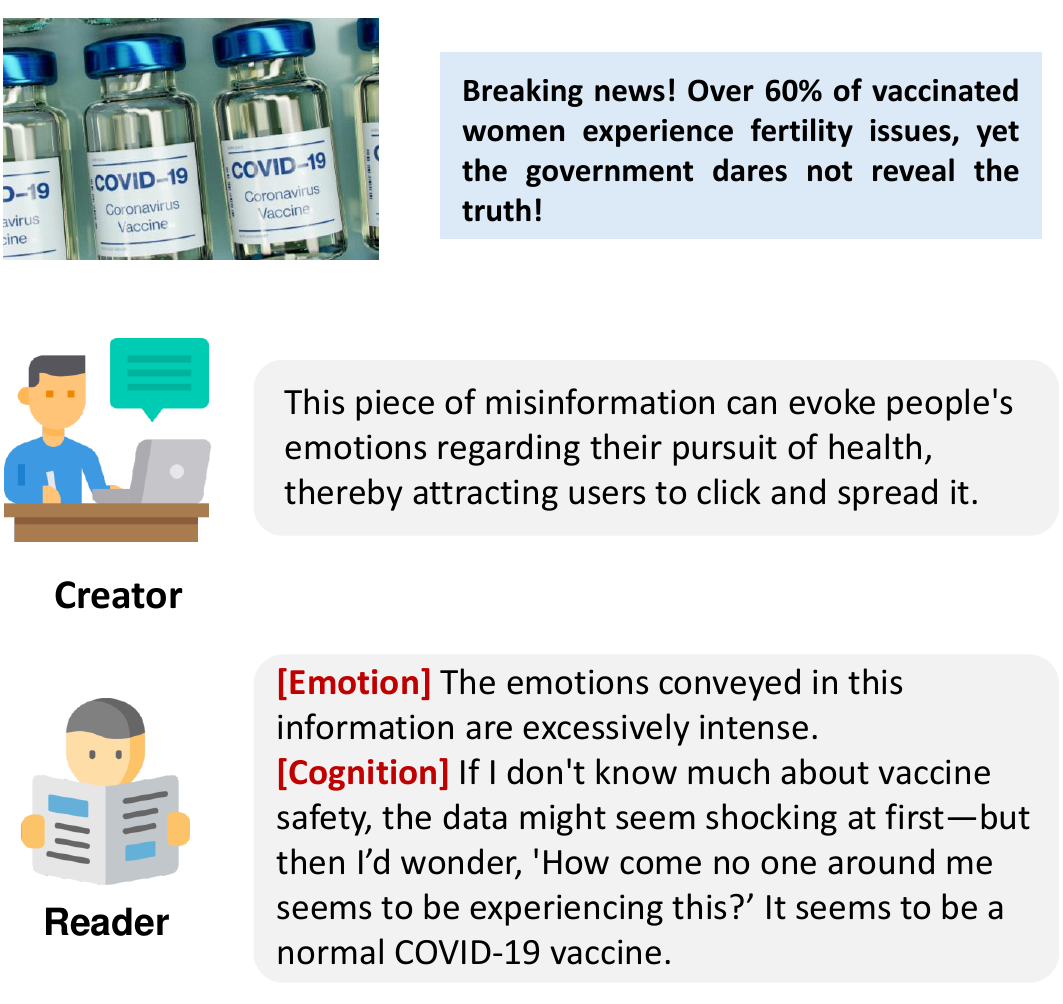}
    \vspace{-1em}
    \caption{An Example of Misinformation.}
    \vspace{-1em}
    \label{fig:mot}
\end{figure}

In recent years, misinformation detection has advanced significantly, with researchers incorporating text analysis~\cite{reddy2020text,bangyal2021detection}, user behavior modeling~\cite{ruchansky2017csi,shu2019role}, image recognition~\cite{steinebach2019fake}, and other technologies~\cite{nan2024let,kim2018leveraging}. Many studies have also explored the role of emotion in misinformation detection, leveraging emotional signals to improve classification accuracy~\cite{liu2024emotion,alonso2021sentiment}. Misinformation spreads not just because of emotions but also due to deeper psychological and cognitive biases. It depends not only on the content itself but also on how people interpret it—shaped by the creators' intent and the reader's biases~\cite{zollo2015emotional,martel2020reliance,bakir2018fake}. As shown in Figure~\ref{fig:mot}, this type of guided misinformation is particularly common on social media in everyday life. The creator aims to influence readers' emotions, guiding them to click and spread the content. Therefore, relying solely on an emotion-based analysis of the content may lead to falling into the creator’s trap and believing that "this is true." However, if both cognitive and emotional perspectives are considered to empathize with both the content and the creator, it becomes much easier to identify this as misinformation.

This phenomenon inspires us to draw from the theory of empathy in psychology to reconstruct the theoretical framework for misinformation detection. Empathy, as a core human ability to understand others' psychological states, encompasses two complementary dimensions: cognitive empathy and emotional empathy~\cite{rankin2005patterns,smith2006cognitive}. Cognitive empathy enables us to adopt others' perspectives, understand their thought processes, belief systems, and behavioral intentions, which is crucial for analyzing the strategies of misinformation creators (focusing on ``what"); while emotional empathy allows us to perceive and experience others' emotional states, which has unique value in simulating the emotional impact of misinformation on audiences (focusing on "how"). Integrating these two empathic abilities in misinformation detection not only reveals the deeper motivations and strategies of creators but also simulates the emotional and cognitive response patterns of different recipients, thereby constructing a more comprehensive and humanized detection framework.

Based on this theory, we propose a dual-aspect empathy framework (DAE), which organically combines cognitive and affective empathy from the dual perspectives of creators and readers, providing a deeper analytical path for multimodal misinformation detection. In the creator dimension, we analyze the characteristics of textual and visual content to obtain the creator's empathy vector; in the reader dimension, we use LLMs to simulate the emotional responses and cognitive judgment processes of diverse reader groups, assess the perceived credibility of information, and carefully design an empathy-aware filtering mechanism to ensure the authenticity and diversity of simulated responses. Then, we incorporate the gap between the two as a feature in our analysis to measure their emotional deviation. This dual-dimensional framework can capture the deep psychological dynamics of misinformation propagation, breaking through the limitations of traditional methods to achieve more precise and explanatory detection effects. 

The main contributions of our work are:

\begin{itemize}
    \item We propose a dual-aspect empathy framework, which is a multi-level analysis based on the perspectives of creators and readers, combining cognitive and emotional empathy features.
    \item In order to obtain the empathy characteristics of different groups of readers, we design a innovative affective empathy data generation and filtering mechanism.
    \item Experiments are conducted on two benchmark datasets,
which demonstrate that our model is a competitive method for multimodal misinformation detection.
\end{itemize}
%!TEX root = ../main.tex
\section{Related Work}
\label{sec:related}

\subsection{Misinformation Detection}
Misinformation refers to false or misleading information that is spread, regardless of intent to deceive~\cite{abdali2024multi}. It poses significant threats to public opinion, social stability, and information security~\cite{zhou2019fake}. In recent years, detection methods have continuously evolved from traditional machine learning approaches to deep learning~\cite{zhou2020survey,broda2024misinformation,alghamdi2024comprehensive}. Early misinformation primarily relied on manual feature engineering, making predictions by extracting textual linguistic features and propagation characteristics combined with shallow classification models~\cite{alnabhan2024fake}. However, these methods have limitations in semantic understanding and contextual analysis, making it difficult to capture subtle expressions and complex contexts~\cite{10308424}.

With the rise of LLMs, models have improved in capturing long-distance semantic associations and logical reasoning\cite{hu2024bad}. Recent works such as Dm-inter~\cite{wang2024misinformation} and JSDRV~\cite{yang2024reinforcement} enhanced detection via reasoning features and data augmentation. Other research leverages user comments as auxiliary information; for example, simulated comments~\cite{nan2024let} and semantic consistency exploration~\cite{wu2023human}.

Multimodal learning has further promoted text-image integration. Models like att-RNN~\cite{jin2017multimodal} and CAFE~\cite{chen2022cross} have advanced feature fusion and modal consistency. In terms of explainability, SNIFFER~\cite{qi2024sniffer} and DELL~\cite{wan2024dell} improved transparency and reliability.

However, the research above overlooks the psychological factors behind misinformation propagation, particularly the emotional manipulation of creators and the affective-cognitive responses of readers. To address this gap, our method uniquely combines LLMs with empathy theory from the perspectives of creators and readers, integrating cognitive and affective empathy analysis to better understand the psychological dynamics of misinformation.

\subsection{Empathy in Social Computing}
Empathy, as the ability to understand and experience others' emotional and cognitive states, is playing an increasingly important role in artificial intelligence research~\cite{perry2023ai,sorin2024large}. Researchers are attempting to integrate empathy capabilities into natural language processing tasks to promote development in application areas such as human-computer dialogue~\cite{alam2018annotating}, and mental health~\cite{lee2023chain}.

In misinformation detection, the emotion and empathy perspective provides a new analytical dimension. ~\citet{ma2024simulated} simulated user responses through the susceptibility to misinformation test, while ~\cite{zubiaga2017exploiting} improved detection performance by mining the relationship between publisher emotions and social emotions. These studies indicate that introducing emotional and empathetic perspectives helps reveal the deep psychological mechanisms of misinformation propagation, providing a powerful complement to traditional technical feature analysis.

In the assessment and enhancement of LLMs' empathy capabilities, comparative studies by~\citet{qian2023harnessing} and~\citet{lee2024large} show that LLMs have demonstrated abilities approaching or even exceeding humans in empathetic dialogue generation. Furthermore, the studies of~\citet{zhu2024empathizing} and~\citet{wang2022empathetic} enhance LLMs' empathy expression capabilities at the algorithm and architecture levels. However, most existing research often focuses only on a single dimension of empathy ability, lacking systematic consideration of the multidimensional nature of empathy.

%!TEX root = ../main.tex

\section{Method}
\label{sec:method}

\begin{figure*}
    \centering
    \includegraphics[width=1\linewidth]{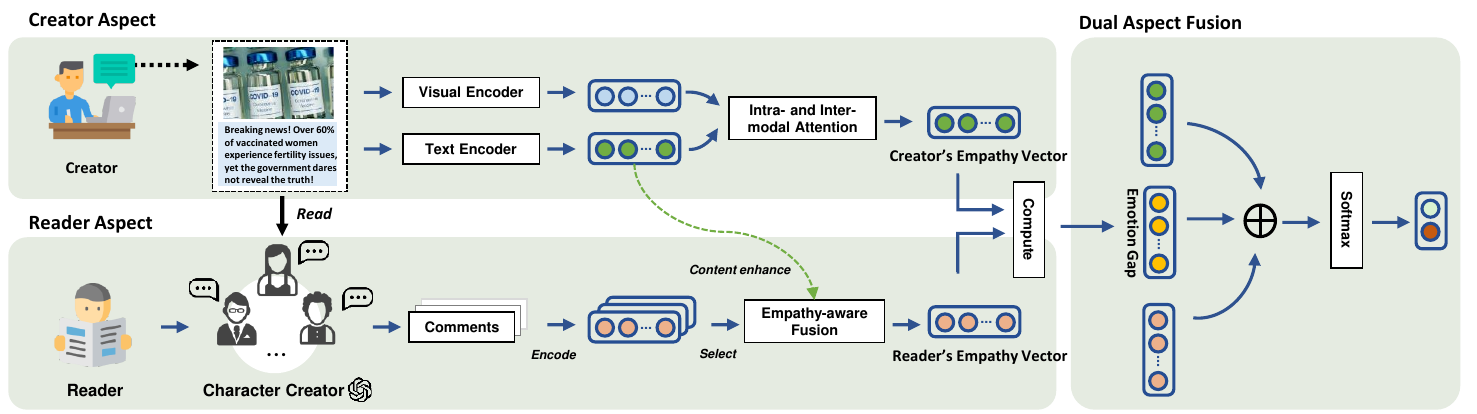}
    \vspace{-1em}
    \caption{Overview of Our Proposed Dual-Aspect Empathy Framework (DAE)}
    \vspace{-1em}
    \label{fig:framework}
\end{figure*}

In this section, we begin with the task definition, followed by an overview of the DAE framework. Each component is then described in detail, as shown in Figure~\ref{fig:framework}.

\subsection{Task Definition}

In this work, we focus specifically on \textbf{news-based} misinformation. The task of multimodal misinformation detection aims to determine the veracity of a given news based on its multimodal input information. Formally, each news is represented as:

\[
D_i = \{ T_i, I_i, C_i \}
\]

where \( D_i \) denotes the multimodal data of the \( i \)-th news, \( T_i \) represents the textual content, \( I_i \) corresponds to the associated image, and \( C_i \) is the set of user comments. The objective of this task is to classify each tweet as either \textit{true} or \textit{false}, making it a binary classification problem:

\[
f(D_i) \to \{ \text{TRUE}, \text{FALSE} \}
\]

where \( f(\cdot) \) represents the classification model that takes multimodal data \( D_i \) as input and predicts whether the tweet contains misinformation.

\subsection{Comment Generation and Filtering}

To capture readers' ``empathic" responses to textual and visual information, we leverage LLMs to simulate readers and automatically generate comments. Referring to the work of ~\citet{nan2024let}, different demographic groups (such as age, gender, and educational background) perceive content with varying levels of understanding and perspectives. These variations provide valuable insights into our study of empathy, including both emotional and cognitive empathy. Therefore, we have designed an LLM-based comment generation and filtering mechanism to comprehensively capture readers' empathetic responses.

\subsubsection{Demographic-Based Comment Generation}

The augmentation process begins with generating comments using LLMs to simulate diverse social perspectives. By instructing LLMs to embody different demographic backgrounds, we create a more comprehensive representation of real-world public discourse. We define a set of user profiles:

\begin{equation}
    \text{Profiles} = \{ P_1, P_2, P_3, \dots, P_n \}
\end{equation}

where each profile \( P_i \) is characterized by:

- \textbf{Gender}: Male/Female.

- \textbf{Age group}: Youth (18-35), Middle-aged (36-65), Elderly (65+).

- \textbf{Education level}: Below Bachelor's, Bachelor's, Postgraduate.

Given a news article \( N \) with text content, associated image \( I \), and any existing comment set \( C = \{ c_1, c_2, \dots, c_m \} \), we prompt GPT-4o to simulate how individuals from different demographic backgrounds would respond:

\begin{equation}
    G = \text{LLM}(N, C, \text{Profiles}, I)
\end{equation}

This LLM-based simulation approach allows us to capture the nuanced ways in which diverse individuals might process, interpret, and respond to potentially misleading multimodal content. The simulation is guided by two fundamental empathic dimensions:

\textbf{Cognitive empathy}: We instruct the LLM to simulate how individuals from different backgrounds would cognitively process the information, including their ability to identify inconsistencies, evaluate credibility based on prior knowledge, and formulate opinions that reflect their educational and experiential background. As the example in Fig.~\ref{fig:mot}, individuals with medical or academic backgrounds may immediately question the validity of a “60\%” claim, seeking evidence and suspecting misinformation. Others, especially with less scientific literacy, may react with initial alarm but later express doubt when the claim contradicts their everyday experience.

\textbf{Emotional empathy}: The LLM simulates emotional responses that might arise from different demographic groups, capturing variations in sentiment intensity, emotional triggers, and affective expressions that correlate with age, gender, and educational differences. As the example in Fig.~\ref{fig:mot}, young women may feel intense anxiety, particularly concerning future family planning. Parents may react with anger or betrayal before turning to verification.

By directing the LLM to embody these empathic dimensions while considering demographic factors, we obtain a rich spectrum of simulated reactions that mirrors the complexity and diversity of human responses to multimodal information.

\subsubsection{Empathy-Driven Comment Filtering}

Following the simulation stage, we implemented a robust filtering mechanism to ensure only high-quality, empathy-driven comments were retained. The refined comment set \( C' \) was obtained as follows:

\begin{equation}
    C' = \text{Filter}(C \cup G)
\end{equation}

The filtering criteria prioritized comments that demonstrated either cognitive or emotional empathy while discarding those that were irrelevant to the news topic, lacked empathy signals, contained offensive or inappropriate content, included social media artifacts (e.g., @mentions, hashtags), or were written in languages other than English (these were translated).

To maintain contextual richness, we enforced a minimum threshold of five high-quality comments per article. For articles with an excessive number of comments (\( |C| > 50 \)), a length-based prioritization was applied, retaining longer comments that typically offer more substantive insights.

\subsection{Feature Extraction and Encoding}

For text \(T_i\) and image \(I_i\) of the original news, we extract features through dedicated encoders:  
\begin{align}
T_i &= \text{Encoder}_{\text{text}}(T_i), \\  
I_i &= \text{Encoder}_{\text{visual}}(I_i),
\end{align}  
where \(T_i \in \mathbb{R}^{L_t \times d_t}\) and \(I_i \in \mathbb{R}^{L_i \times d_i}\). To ensure feature space consistency, we map textual and visual features to a unified dimensional space through linear projection:  
\begin{align}
T_i' &= W_t T_i, \\  
I_i' &= W_i I_i,
\end{align}  
where \(T_i' \in \mathbb{R}^{L_t \times d}\) and \(I_i' \in \mathbb{R}^{L_i \times d}\). Unlike traditional methods that focus solely on content itself, we specifically introduce user comments as an important supplement to the reader's perspective. For the relevant comment collection \(C_i\), we extract semantic representation for each comment:  
\begin{align}
C_{ij} &= \text{Encoder}_{\text{text}}(C_{ij})_{[\text{CLS}]},
\end{align}  
where \([\text{CLS}]\) represents the classification token vector output by RoBERTa, and \(C_{ij} \in \mathbb{R}^{d_t}\).  All comments form the collective representation:  
\begin{align}
C_i &= [C_{i1}; C_{i2}; \ldots; C_{iN}],
\end{align}  
where \(C_i \in \mathbb{R}^{N \times d_t}\). Similarly, we apply linear dimensionality reduction to comment features:  
\begin{align}
C_i' &= W_c C_i,
\end{align}  
with \(C_i' \in \mathbb{R}^{N \times d}\).

\subsection{Feature Enhancement and Interaction}

To capture sequential internal dependencies, we apply multi-head self-attention mechanism (MHSA) to text, image, and comment features respectively:
\begin{equation}
T_i'' = \text{MHSA}(T_i'), I_i'' = \text{MHSA}(I_i'), C_i'' = \text{MHSA}(C_i')
\end{equation}

Through cross-modal attention mechanism, we fuse textual and visual features:
\begin{align}
TI_i &= \text{CrossAttn}(T_i'', I_i'')
\end{align}

For the comment collection, we adopt an adaptive filtering mechanism based on semantic importance:
\begin{align}
C_i^{\text{k}} &= \text{top-k}(C_i'', k)
\end{align}
This ensures that the selected comments maximally reflect the public's genuine reactions to the information, providing richer reader perspective evidence.

\subsection{Dual-Aspect Empathy Feature Construction}

\subsubsection{Basic Feature Extraction}

We begin by extracting two fundamental types of features using pooling operations. Specifically, we apply mean pooling to obtain the creator features, reflecting the overall intention and cognitive patterns of the information producer. In contrast, max pooling is employed to derive the reader features, capturing the most salient reactions from the audience:
\begin{equation}
e_{\text{creator}} = \text{MeanPooling}(T_i''), 
e_{\text{reader}} = \text{MaxPooling}(C_i^{\text{k}})
\end{equation}

\subsubsection{Empathy-aware Fusion}

To model the cognitive aspect of empathy, we integrate the extracted creator and reader features with multimodal representations derived from both text and images. This results in a cognitively informed fusion representation:
\begin{align}
h_c &= \text{Concat}(e_{\text{creator}}, e_{\text{reader}}, TI_i)
\end{align}

This fusion enables the model to jointly capture the creator’s communicative intentions, the reader’s interpretative stance, and the semantic alignment (or misalignment) between textual and visual modalities. It reflects a holistic understanding of how information is constructed and perceived.

\subsubsection{Emotion Gap Computing}

Moving beyond factual analysis, we introduce an \textbf{emotional empathy} dimension to better understand the affective manipulation mechanisms commonly found in misinformation. The first step involves computing the emotional discrepancy between the creator and the reader:
\begin{align}
e_{\text{gap}} &= |e_{\text{creator}} - e_{\text{reader}}|
\end{align}

This \textit{emotion gap} captures key signals of emotional incitement, where creators may intentionally exaggerate or distort emotional cues to influence reader perception. We then construct a comprehensive emotional empathy representation by fusing the emotional signals of both parties along with their differences:
\begin{align}
h_e &= \text{Concat}(e_{\text{creator}}, e_{\text{reader}}, e_{\text{gap}})
\end{align}

\subsection{Fusion and Classification}

To form a unified representation, we concatenate the cognitive and emotional empathy features:
\begin{align}
h_f &= \text{Concat}(h_c, h_e)
\end{align}

This final representation is passed through a multilayer perceptron (MLP) followed by a Softmax activation to predict the veracity of the content:
\begin{align}
\hat{y}_i &= \text{Softmax}(\text{MLP}(h_f))
\end{align}

For training, we employ a cross-entropy loss function with label smoothing, along with $L_2$ regularization to prevent overfitting:
\begin{align}
L(\theta) &= -\sum_i \sum_c y_{ic} \log \hat{y}_{ic} + \lambda \|\theta\|^2
\end{align}
where $y_{ic}$ denotes the true label distribution, and $\lambda$ is the regularization coefficient.

%!TEX root = ../main.tex

\section{Experiment}
\label{sec:expt}

\subsection{Experimental Setup}

\textbf{Datasets.} We evaluate model performance on two widely used datasets: PHEME~\cite{zhang2024natural} and PolitiFact~\cite{jin2022towards}. The PHEME dataset is a multimodal misinformation collection that includes five breaking events: Charlie Hebdo, Ferguson, Germanwings Crash, Ottawa Shooting, and Sydney Siege. We classify tweets, images, and comments related to each news item. PolitiFact is also a multimodal fake news detection dataset. The distribution of true and false news in both datasets is shown in the table, and we adopt an $8:2$ ratio for the training and test set split.

\begin{table}[htbp]
    \centering
    \vspace{-1em}
    \caption{Partition Datasets of Two Datasets}
        \vspace{-1em}
    \begin{tabular}{lccc}
        \toprule
        \multirow{2}{*}{Partition} & \multicolumn{2}{c}{Datasets} \\
        \cmidrule(lr){2-3}
        & PHEME & PolitiFact \\
        \midrule
        \# of Fake News & 1972 & 432 \\
        \# of True News & 3830 & 624 \\
        \# of Images & 3670 & 783 \\
        \bottomrule
    \end{tabular}
    \vspace{-1em}
\end{table}

\textbf{Evaluation Metrics.} We use accuracy, precision, recall, and F1-score as evaluation metrics.

\textbf{Experimental Environment.} In our experiments, the maximum text sequence length is set to $512$, and the maximum length for each comment sequence is $128$. As the number of comments associated with news items varies, we limit each news item to a maximum of $15$ real comments and introduce GPT-generated supplementary comments. For text encoding, we use the pre-trained RoBERTa-base model, while visual encoding employs the pre-trained Swin-base model, freezing the first $n-1$ layers of the Swin Transformer and only training the last layer. The multi-head self-attention network uses $8$ attention heads. Image features are extracted through the Swin Transformer, with output sequence length determined by the Swin model's settings. In the comment selection mechanism, the Pointer Network selects the top-k comments, where $k=5$. Additionally, the initial learning rate is $0.001$, the optimizer is Adam, and the loss function is cross-entropy loss.

\subsection{Baselines}

To verify the performance of our model, we selected several competitive misinformation detection methods as baselines. 
For PHEME dataset, the baselines are MVAE~\cite{khattar2019mvae}, SAFE~\cite{zhou2020similarity}, SpotFake~\cite{singhal2019spotfake}, CAFE~\cite{chen2022cross}, MCAN~\cite{wu2021multimodal}, KDIN~\cite{sun2023inconsistent}, LIIMR~\cite{singhal2022leveraging}, BMR~\cite{ying2023bootstrapping}, while the baselines for PolitiFact are XLNet~\cite{yang2019xlnet}, GCAN~\cite{ma2016detecting}, SpotFake~\cite{singhal2019spotfake}, KAN~\cite{dun2021kan}, MCAN~\cite{wu2021multimodal}, MCAN-A~\cite{wu2021multimodal}, SpotFake+~\cite{singhal2020spotfake+}, QMFND~\cite{qu2024qmfnd}.

\begin{table}[t!]
\centering
\caption{Main Experiment Result}
\begin{tabular}{llcccc}
\toprule
\textbf{Dataset} & \textbf{Method} & \textbf{Acc(\%)} & \textbf{P.(\%)} & \textbf{R.(\%)} & \textbf{F1(\%)} \\
\midrule
\multirow{9}{*}{\textbf{PHEME}}
& MVAE         & 77.6  & 73.5  & 72.3  & 72.8 \\
& SAFE         & 80.7  & 78.7  & 78.9  & 79.1 \\
& SpotFake     & 84.5  & 80.9  & 83.6  & 82.2 \\
& CAFE         & 83.2  & 79.6  & 79.4  & 79.5 \\
& MCAN         & 86.1  & 83.0  & 84.0  & 83.5 \\
& KDIN         & 84.6  & 81.5  & 80.4  & 80.9 \\
& LIIMR        & 87.0  & 84.8  & 83.1  & 83.9 \\
& BMR          & 88.4  & 87.2  & 84.0  & 85.5 \\
& \textbf{DAE (Ours)} & \textbf{89.8} & \textbf{89.9} & \textbf{89.8} & \textbf{89.8} \\
\midrule
\multirow{9}{*}{\textbf{PolitiFact}}
& XLNet        & 84.7 & 90.5 & 70.4 & 79.2 \\
& GCAN         & 80.8 & 79.5 & 84.2 & 83.5 \\
& SpotFake     & 77.9 & 81.5 & 69.3 & 74.9 \\
& KAN          & 85.9 & 86.9 & 85.0 & 85.4 \\
& MCAN         & 84.6 & 85.1 & 82.9 & 84.0 \\
& MCAN-A       & 80.9 & 82.7 & 75.8 & 79.2 \\
& SpotFake+    & 78.9 & 80.3 & 75.3 & 77.7 \\
& QMFND        & 84.6 & \textbf{92.7} & 85.3 & 88.8 \\
& \textbf{DAE (Ours)} & \textbf{90.6} & 90.8 & \textbf{90.6} & \textbf{90.4} \\
\bottomrule
\end{tabular}
\vspace{-1em}
\label{table:main_result}
\end{table}

\begin{figure}[t!]
    \centering
    \includegraphics[width=1\linewidth]{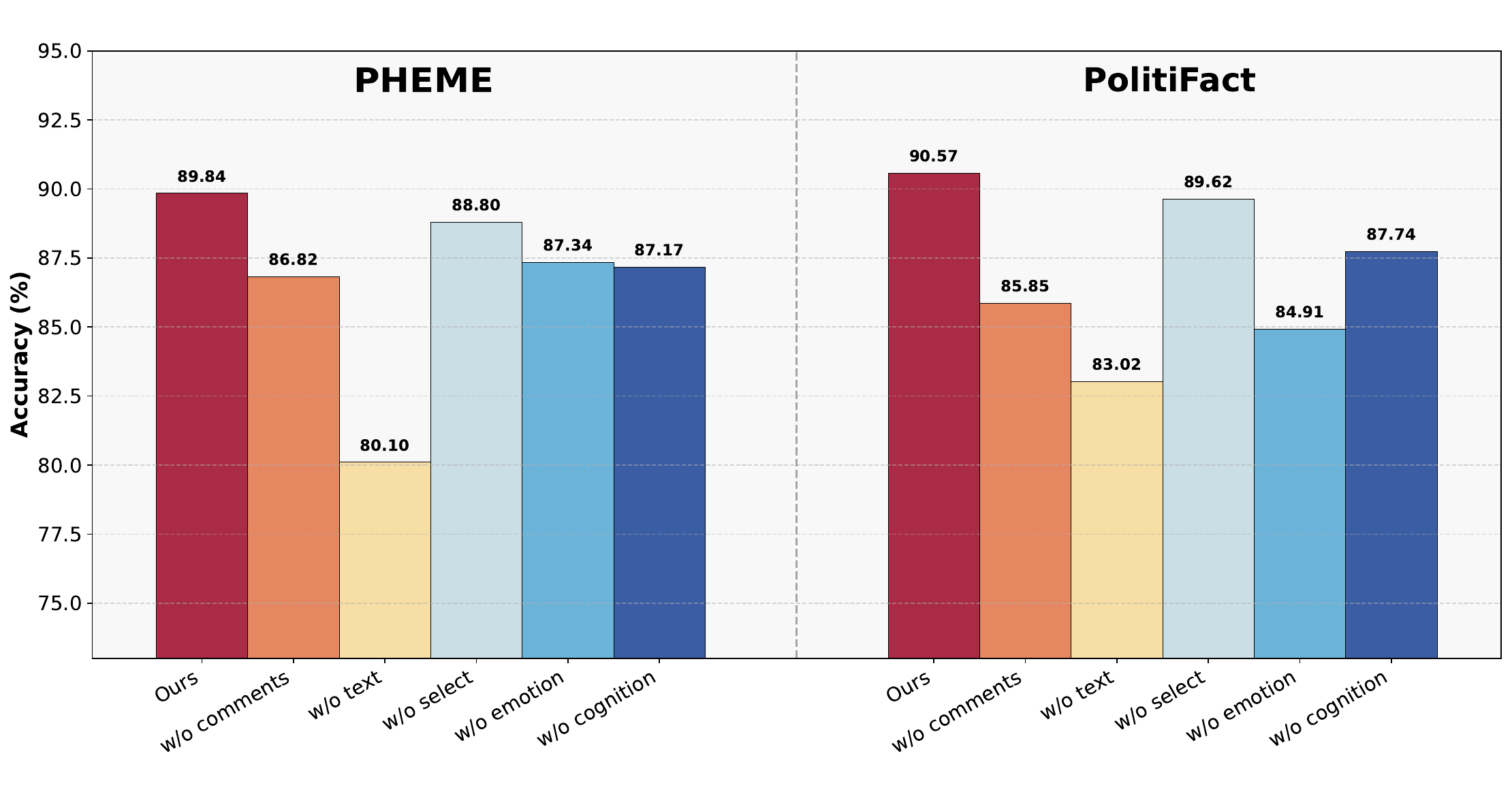}
    \vspace{-2em}
    \caption{Results of Ablation Study}
    \label{fig:ablation}
    \vspace{-2em}
\end{figure}

\subsection{Main Experiment}

To validate the effectiveness of our approach, we conducted experiments on two datasets. The main experiment results are shown in Table~\ref{table:main_result}. Overall, our method outperforms the baselines on both datasets.

Specifically, on the PHEME dataset, our proposed DAE model achieves an accuracy of 89.8\%, significantly surpassing the state-of-the-art BMR (88.4\%) and LIIMR (87.0\%). In detail, BMR leverages a bidirectional multimodal reasoning mechanism to deeply capture semantic interactions between textual and visual information, while LIIMR employs implicit interaction inference to explore latent relationships between multimodal features. However, both methods mainly focus on the content itself and fail to explicitly consider users' feedback in emotional and cognitive dimensions. In contrast, DAE not only integrates multimodal information, including text, images, and comments, but also explicitly models users' emotional resonance and cognitive skepticism through a dual-dimensional empathy mechanism. This comprehensive modeling better reflects users' real feedback on news, enabling the model to more accurately identify the authenticity of news.

On the PolitiFact dataset, DAE also achieves a significant improvement, with an accuracy of 90.6\%, noticeably outperforming KAN (85.9\%), which employs a kernel-based attention mechanism, and QMFND (84.6\%), which integrates image and text features using a quantum convolutional neural network. KAN primarily measures textual semantic similarity through kernel-based attention, capturing rich semantic information but neglecting the interactive relationship between user comments and news content. QMFND, despite fusing textual and visual features, adopts a static feature extraction approach that fails to fully exploit the emotional and cognitive cues embedded in user feedback. In contrast, DAE incorporates a comment selection mechanism to explicitly capture users' cognitive consistency and emotional divergence regarding news content. This allows the model to deeply explore users' resonance or skepticism toward the news.

\begin{figure*}[t!]
    \centering
    \includegraphics[width=1\linewidth]{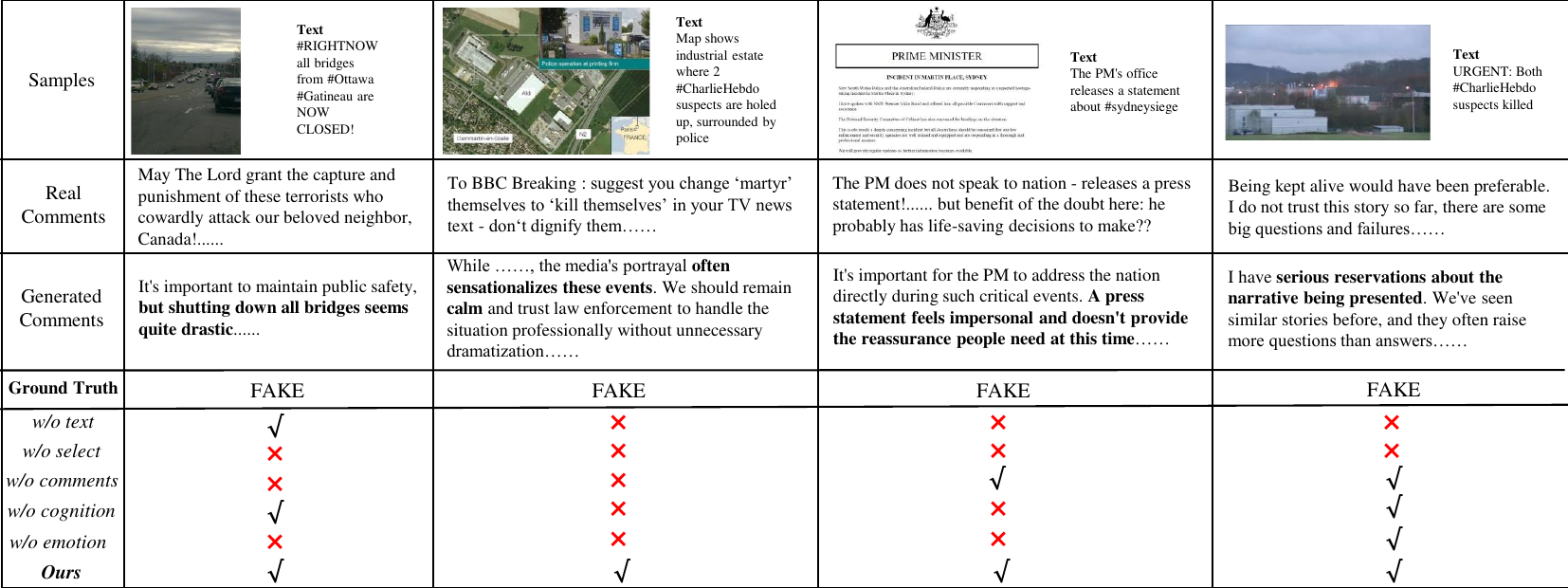}
    \vspace{-1em}
    \caption{Several Cases in the Test Set of Datasets}
    \vspace{-1em}
    \label{fig:case_study}
\end{figure*}

\begin{figure}[t!]
    \centering
    \includegraphics[width=1\linewidth]{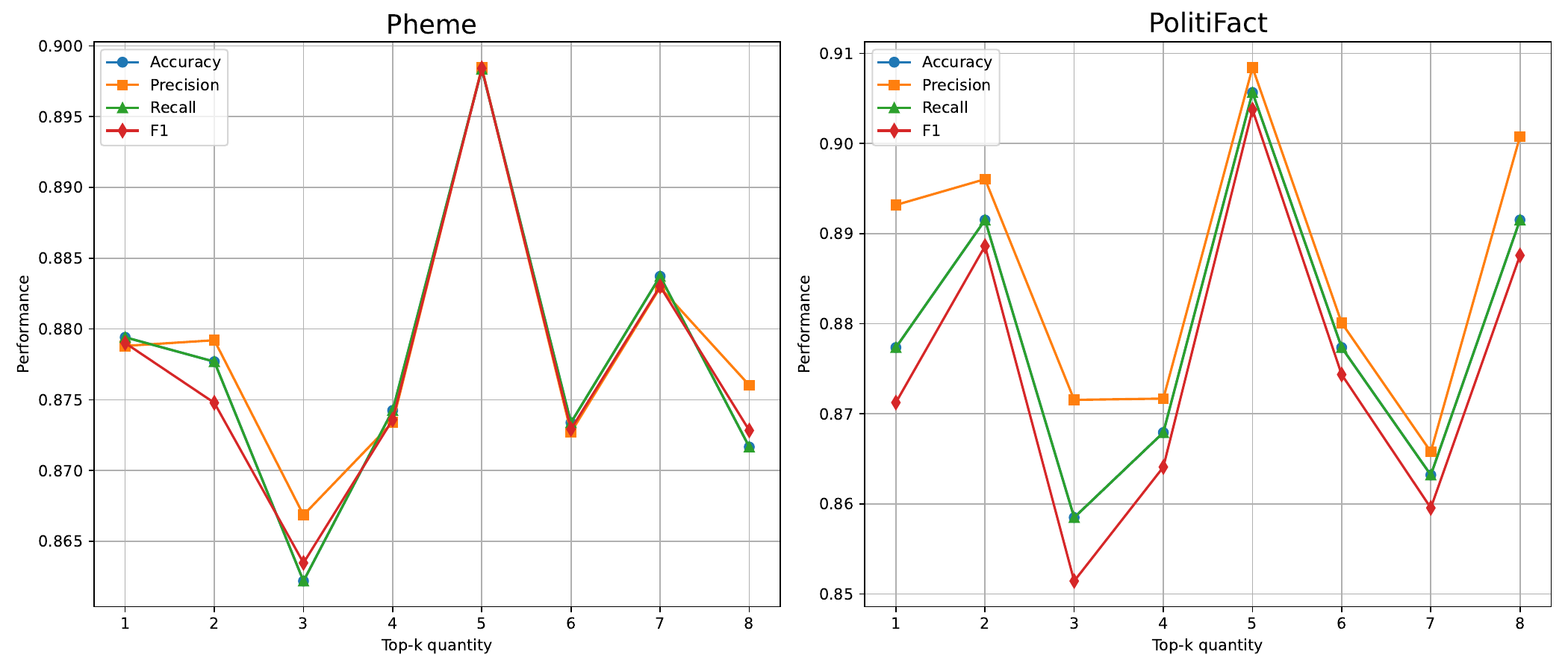}
    \vspace{-1em}
    \caption{Results of Parameter Analysis}
    \label{fig:parameter}
    \vspace{-2em}
\end{figure}

\subsection{Ablation Study}

We verified the effectiveness of each component in the dual-dimension empathy framework through five ablation settings, including: 
\begin{itemize}
    \item $w/o$ comments: DAE with removing reader comments input.
    \item $w/o$ text: DAE with removing tweet text input.
    \item $w/o$ select: DAE with removing comment selection mechanism.
    \item $w/o$ emotion: DAE with removing emotional empathy.
    \item $w/o$ cognition: DAE with removing cognitive empathy.
\end{itemize}

The experimental results show that tweet text is fundamental, as its removal leads to the largest accuracy drop (PHEME: from 89.84\% to 80.10\%; PolitiFact: from 90.57\% to 83.02\%). Removing comments also degrades performance (PHEME: -3.02\%, PolitiFact: -4.72\%), confirming their complementary role.

The comment selection mechanism significantly improves accuracy, as its removal reduces performance (PHEME: -1.04\%, PolitiFact: -0.95\%), verifying the effectiveness of top-k filtering.

Emotional and cognitive empathy impact datasets differently. On PolitiFact, removing emotional empathy leads to a larger drop (from 90.57\% to 84.91\%) than cognitive empathy (from 90.57\% to 87.74\%). Conversely, on PHEME, cognitive empathy has a greater effect (from 89.84\% to 87.17\%) than emotional empathy (from 89.84\% to 87.34\%).

These results highlight the necessity of tweet text, reader comments, comment selection, and the dual-dimension empathy mechanism in model performance.

\subsection{Parameter Analysis}

In the parameter experiments, we focused on examining the selection quantity in the top-k comment selection mechanism. We tested different comment selection quantities from 1 to 8 on both PHEME and PolitiFact datasets, with results shown in the figure. The experiments found that when the number of selected comments is too few (such as 1 or 2), the model cannot obtain sufficient information to support decision-making; when the number of selected comments is too many (such as 7 or 8), the model is easily disturbed by irrelevant or redundant comments, and performance also declines.

In both datasets, the model achieves optimal performance when the comment selection quantity is set to 5 (accuracy of 89.8\% on the PHEME dataset and 90.6\% on the PolitiFact dataset). This indicates that a moderate number of selected comments can more effectively help the model focus on truly valuable comment information, thereby enhancing the accuracy of fake information detection.

\subsection{Case Study}

As shown in Fig.~\ref{fig:case_study}, we further verified the effectiveness of the dual-dimension empathy framework and the importance of each module. Our findings reveal that emotional and cognitive empathy play a crucial role in assessing news authenticity.

In the first case, genuine comments conveyed anger, while AI-generated comments remained rational yet critical. Removing emotional empathy significantly impaired the model’s accuracy, highlighting its importance. In the second case, genuine comments were more aggressive, whereas AI-generated ones promoted rationality by critiquing media exaggeration. The removal of cognitive empathy notably affected the model’s judgment, underscoring its significance. In the third case, genuine comments criticized the government’s response, while AI-generated ones emphasized direct communication. The absence of user comments led to a sharp decline in detection accuracy, proving their essential role. In the final case, genuine comments expressed distrust, while AI-generated ones questioned the official narrative. Removing both cognitive and emotional empathy caused significant misjudgments, reaffirming their combined effect.

Overall, our study provides strong empirical evidence for the indispensable role of emotional and cognitive empathy in detecting misinformation, demonstrating their theoretical interdependence. 
%!TEX root = ../main.tex
\section{Conclusion}
\label{sec:conclusion}

In conclusion, we introduce a dual-aspect empathy framework for multimodal misinformation detection, which integrates cognitive and affective empathy to provide a deeper, more human-centered approach. By analyzing the psychological strategies of creators and simulating readers' emotional and cognitive responses, the framework achieves more precise and explanatory detection compared to traditional methods. 

Future directions can focus on real-time empathy modeling to adapt to evolving misinformation tactics and extending the framework to cross-cultural contexts for greater robustness.

\bibliographystyle{ACM-Reference-Format}
\bibliography{sample-base}


\begin{thebibliography}{60}

%%% the Web address.

\ifx \showCODEN    \undefined \def \showCODEN     #1{\unskip}     \fi
\ifx \showISBNx    \undefined \def \showISBNx     #1{\unskip}     \fi
\ifx \showISBNxiii \undefined \def \showISBNxiii  #1{\unskip}     \fi
\ifx \showISSN     \undefined \def \showISSN      #1{\unskip}     \fi
\ifx \showLCCN     \undefined \def \showLCCN      #1{\unskip}     \fi
\ifx \shownote     \undefined \def \shownote      #1{#1}          \fi
\ifx \showarticletitle \undefined \def \showarticletitle #1{#1}   \fi
\ifx \showURL      \undefined \def \showURL       {\relax}        \fi
% invisible to TeX
\providecommand\bibfield[2]{#2}
\providecommand\bibinfo[2]{#2}
\providecommand\natexlab[1]{#1}
\providecommand\showeprint[2][]{arXiv:#2}

\bibitem[Abdali et~al\mbox{.}(2024)]%
        {abdali2024multi}
\bibfield{author}{\bibinfo{person}{Sara Abdali}, \bibinfo{person}{Sina Shaham}, {and} \bibinfo{person}{Bhaskar Krishnamachari}.} \bibinfo{year}{2024}\natexlab{}.
\newblock \showarticletitle{Multi-modal misinformation detection: Approaches, challenges and opportunities}.
\newblock \bibinfo{journal}{\emph{Comput. Surveys}} \bibinfo{volume}{57}, \bibinfo{number}{3} (\bibinfo{year}{2024}), \bibinfo{pages}{1--29}.
\newblock


\bibitem[Alam et~al\mbox{.}(2018)]%
        {alam2018annotating}
\bibfield{author}{\bibinfo{person}{Firoj Alam}, \bibinfo{person}{Morena Danieli}, {and} \bibinfo{person}{Giuseppe Riccardi}.} \bibinfo{year}{2018}\natexlab{}.
\newblock \showarticletitle{Annotating and modeling empathy in spoken conversations}.
\newblock \bibinfo{journal}{\emph{Computer Speech \& Language}}  \bibinfo{volume}{50} (\bibinfo{year}{2018}), \bibinfo{pages}{40--61}.
\newblock


\bibitem[Alghamdi et~al\mbox{.}(2024)]%
        {alghamdi2024comprehensive}
\bibfield{author}{\bibinfo{person}{Jawaher Alghamdi}, \bibinfo{person}{Suhuai Luo}, {and} \bibinfo{person}{Yuqing Lin}.} \bibinfo{year}{2024}\natexlab{}.
\newblock \showarticletitle{A comprehensive survey on machine learning approaches for fake news detection}.
\newblock \bibinfo{journal}{\emph{Multimedia Tools and Applications}} \bibinfo{volume}{83}, \bibinfo{number}{17} (\bibinfo{year}{2024}), \bibinfo{pages}{51009--51067}.
\newblock


\bibitem[Alnabhan and Branco(2024)]%
        {alnabhan2024fake}
\bibfield{author}{\bibinfo{person}{Mohammad~Q Alnabhan} {and} \bibinfo{person}{Paula Branco}.} \bibinfo{year}{2024}\natexlab{}.
\newblock \showarticletitle{Fake news detection using deep learning: A systematic literature review}.
\newblock \bibinfo{journal}{\emph{IEEE Access}} (\bibinfo{year}{2024}).
\newblock


\bibitem[Alonso et~al\mbox{.}(2021)]%
        {alonso2021sentiment}
\bibfield{author}{\bibinfo{person}{Miguel~A Alonso}, \bibinfo{person}{David Vilares}, \bibinfo{person}{Carlos G{\'o}mez-Rodr{\'\i}guez}, {and} \bibinfo{person}{Jes{\'u}s Vilares}.} \bibinfo{year}{2021}\natexlab{}.
\newblock \showarticletitle{Sentiment analysis for fake news detection}.
\newblock \bibinfo{journal}{\emph{Electronics}} \bibinfo{volume}{10}, \bibinfo{number}{11} (\bibinfo{year}{2021}), \bibinfo{pages}{1348}.
\newblock


\bibitem[Bakir and McStay(2018)]%
        {bakir2018fake}
\bibfield{author}{\bibinfo{person}{Vian Bakir} {and} \bibinfo{person}{Andrew McStay}.} \bibinfo{year}{2018}\natexlab{}.
\newblock \showarticletitle{Fake news and the economy of emotions: Problems, causes, solutions}.
\newblock \bibinfo{journal}{\emph{Digital journalism}} \bibinfo{volume}{6}, \bibinfo{number}{2} (\bibinfo{year}{2018}), \bibinfo{pages}{154--175}.
\newblock


\bibitem[Bangyal et~al\mbox{.}(2021)]%
        {bangyal2021detection}
\bibfield{author}{\bibinfo{person}{Waqas~Haider Bangyal}, \bibinfo{person}{Rukhma Qasim}, \bibinfo{person}{Najeeb~Ur Rehman}, \bibinfo{person}{Zeeshan Ahmad}, \bibinfo{person}{Hafsa Dar}, \bibinfo{person}{Laiqa Rukhsar}, \bibinfo{person}{Zahra Aman}, {and} \bibinfo{person}{Jamil Ahmad}.} \bibinfo{year}{2021}\natexlab{}.
\newblock \showarticletitle{Detection of Fake News Text Classification on COVID-19 Using Deep Learning Approaches}.
\newblock \bibinfo{journal}{\emph{Computational and mathematical methods in medicine}} \bibinfo{volume}{2021}, \bibinfo{number}{1} (\bibinfo{year}{2021}), \bibinfo{pages}{5514220}.
\newblock


\bibitem[Broda and Str{\"o}mb{\"a}ck(2024)]%
        {broda2024misinformation}
\bibfield{author}{\bibinfo{person}{Elena Broda} {and} \bibinfo{person}{Jesper Str{\"o}mb{\"a}ck}.} \bibinfo{year}{2024}\natexlab{}.
\newblock \showarticletitle{Misinformation, disinformation, and fake news: lessons from an interdisciplinary, systematic literature review}.
\newblock \bibinfo{journal}{\emph{Annals of the International Communication Association}} \bibinfo{volume}{48}, \bibinfo{number}{2} (\bibinfo{year}{2024}), \bibinfo{pages}{139--166}.
\newblock


\bibitem[Chen et~al\mbox{.}(2023)]%
        {chen2023spread}
\bibfield{author}{\bibinfo{person}{Sijing Chen}, \bibinfo{person}{Lu Xiao}, {and} \bibinfo{person}{Akit Kumar}.} \bibinfo{year}{2023}\natexlab{}.
\newblock \showarticletitle{Spread of misinformation on social media: What contributes to it and how to combat it}.
\newblock \bibinfo{journal}{\emph{Computers in Human Behavior}}  \bibinfo{volume}{141} (\bibinfo{year}{2023}), \bibinfo{pages}{107643}.
\newblock


\bibitem[Chen et~al\mbox{.}(2022)]%
        {chen2022cross}
\bibfield{author}{\bibinfo{person}{Yixuan Chen}, \bibinfo{person}{Dongsheng Li}, \bibinfo{person}{Peng Zhang}, \bibinfo{person}{Jie Sui}, \bibinfo{person}{Qin Lv}, \bibinfo{person}{Lu Tun}, {and} \bibinfo{person}{Li Shang}.} \bibinfo{year}{2022}\natexlab{}.
\newblock \showarticletitle{Cross-modal ambiguity learning for multimodal fake news detection}. In \bibinfo{booktitle}{\emph{Proceedings of the ACM web conference 2022}}. \bibinfo{pages}{2897--2905}.
\newblock


\bibitem[Dun et~al\mbox{.}(2021)]%
        {dun2021kan}
\bibfield{author}{\bibinfo{person}{Yaqian Dun}, \bibinfo{person}{Kefei Tu}, \bibinfo{person}{Chen Chen}, \bibinfo{person}{Chunyan Hou}, {and} \bibinfo{person}{Xiaojie Yuan}.} \bibinfo{year}{2021}\natexlab{}.
\newblock \showarticletitle{Kan: Knowledge-aware attention network for fake news detection}. In \bibinfo{booktitle}{\emph{Proceedings of the AAAI conference on artificial intelligence}}, Vol.~\bibinfo{volume}{35}. \bibinfo{pages}{81--89}.
\newblock


\bibitem[Guo et~al\mbox{.}(2019)]%
        {guo2019future}
\bibfield{author}{\bibinfo{person}{Bin Guo}, \bibinfo{person}{Yasan Ding}, \bibinfo{person}{Lina Yao}, \bibinfo{person}{Yunji Liang}, {and} \bibinfo{person}{Zhiwen Yu}.} \bibinfo{year}{2019}\natexlab{}.
\newblock \showarticletitle{The future of misinformation detection: new perspectives and trends}.
\newblock \bibinfo{journal}{\emph{arXiv preprint arXiv:1909.03654}} (\bibinfo{year}{2019}).
\newblock


\bibitem[Hu et~al\mbox{.}(2024)]%
        {hu2024bad}
\bibfield{author}{\bibinfo{person}{Beizhe Hu}, \bibinfo{person}{Qiang Sheng}, \bibinfo{person}{Juan Cao}, \bibinfo{person}{Yuhui Shi}, \bibinfo{person}{Yang Li}, \bibinfo{person}{Danding Wang}, {and} \bibinfo{person}{Peng Qi}.} \bibinfo{year}{2024}\natexlab{}.
\newblock \showarticletitle{Bad actor, good advisor: Exploring the role of large language models in fake news detection}. In \bibinfo{booktitle}{\emph{Proceedings of the AAAI Conference on Artificial Intelligence}}, Vol.~\bibinfo{volume}{38}. \bibinfo{pages}{22105--22113}.
\newblock


\bibitem[Jin et~al\mbox{.}(2022)]%
        {jin2022towards}
\bibfield{author}{\bibinfo{person}{Yiqiao Jin}, \bibinfo{person}{Xiting Wang}, \bibinfo{person}{Ruichao Yang}, \bibinfo{person}{Yizhou Sun}, \bibinfo{person}{Wei Wang}, \bibinfo{person}{Hao Liao}, {and} \bibinfo{person}{Xing Xie}.} \bibinfo{year}{2022}\natexlab{}.
\newblock \showarticletitle{Towards fine-grained reasoning for fake news detection}. In \bibinfo{booktitle}{\emph{Proceedings of the AAAI Conference on Artificial Intelligence}}, Vol.~\bibinfo{volume}{36}. \bibinfo{pages}{5746--5754}.
\newblock


\bibitem[Jin et~al\mbox{.}(2017)]%
        {jin2017multimodal}
\bibfield{author}{\bibinfo{person}{Zhiwei Jin}, \bibinfo{person}{Juan Cao}, \bibinfo{person}{Han Guo}, \bibinfo{person}{Yongdong Zhang}, {and} \bibinfo{person}{Jiebo Luo}.} \bibinfo{year}{2017}\natexlab{}.
\newblock \showarticletitle{Multimodal fusion with recurrent neural networks for rumor detection on microblogs}. In \bibinfo{booktitle}{\emph{Proceedings of the 25th ACM international conference on Multimedia}}. \bibinfo{pages}{795--816}.
\newblock


\bibitem[Khattar et~al\mbox{.}(2019)]%
        {khattar2019mvae}
\bibfield{author}{\bibinfo{person}{Dhruv Khattar}, \bibinfo{person}{Jaipal~Singh Goud}, \bibinfo{person}{Manish Gupta}, {and} \bibinfo{person}{Vasudeva Varma}.} \bibinfo{year}{2019}\natexlab{}.
\newblock \showarticletitle{Mvae: Multimodal variational autoencoder for fake news detection}. In \bibinfo{booktitle}{\emph{The world wide web conference}}. \bibinfo{pages}{2915--2921}.
\newblock


\bibitem[Kim et~al\mbox{.}(2018)]%
        {kim2018leveraging}
\bibfield{author}{\bibinfo{person}{Jooyeon Kim}, \bibinfo{person}{Behzad Tabibian}, \bibinfo{person}{Alice Oh}, \bibinfo{person}{Bernhard Sch{\"o}lkopf}, {and} \bibinfo{person}{Manuel Gomez-Rodriguez}.} \bibinfo{year}{2018}\natexlab{}.
\newblock \showarticletitle{Leveraging the crowd to detect and reduce the spread of fake news and misinformation}. In \bibinfo{booktitle}{\emph{Proceedings of the eleventh ACM international conference on web search and data mining}}. \bibinfo{pages}{324--332}.
\newblock


\bibitem[Lee et~al\mbox{.}(2023)]%
        {lee2023chain}
\bibfield{author}{\bibinfo{person}{Yoon~Kyung Lee}, \bibinfo{person}{Inju Lee}, \bibinfo{person}{Minjung Shin}, \bibinfo{person}{Seoyeon Bae}, {and} \bibinfo{person}{Sowon Hahn}.} \bibinfo{year}{2023}\natexlab{}.
\newblock \showarticletitle{Chain of empathy: Enhancing empathetic response of large language models based on psychotherapy models}.
\newblock \bibinfo{journal}{\emph{arXiv preprint arXiv:2311.04915}} (\bibinfo{year}{2023}).
\newblock


\bibitem[Lee et~al\mbox{.}(2024)]%
        {lee2024large}
\bibfield{author}{\bibinfo{person}{Yoon~Kyung Lee}, \bibinfo{person}{Jina Suh}, \bibinfo{person}{Hongli Zhan}, \bibinfo{person}{Junyi~Jessy Li}, {and} \bibinfo{person}{Desmond~C Ong}.} \bibinfo{year}{2024}\natexlab{}.
\newblock \showarticletitle{Large language models produce responses perceived to be empathic}.
\newblock \bibinfo{journal}{\emph{arXiv preprint arXiv:2403.18148}} (\bibinfo{year}{2024}).
\newblock


\bibitem[Liu et~al\mbox{.}(2024)]%
        {liu2024emotion}
\bibfield{author}{\bibinfo{person}{Zhiwei Liu}, \bibinfo{person}{Tianlin Zhang}, \bibinfo{person}{Kailai Yang}, \bibinfo{person}{Paul Thompson}, \bibinfo{person}{Zeping Yu}, {and} \bibinfo{person}{Sophia Ananiadou}.} \bibinfo{year}{2024}\natexlab{}.
\newblock \showarticletitle{Emotion detection for misinformation: A review}.
\newblock \bibinfo{journal}{\emph{Information Fusion}}  \bibinfo{volume}{107} (\bibinfo{year}{2024}), \bibinfo{pages}{102300}.
\newblock


\bibitem[Ma et~al\mbox{.}(2016)]%
        {ma2016detecting}
\bibfield{author}{\bibinfo{person}{Jing Ma}, \bibinfo{person}{Wei Gao}, \bibinfo{person}{Prasenjit Mitra}, \bibinfo{person}{Sejeong Kwon}, \bibinfo{person}{Bernard~J Jansen}, \bibinfo{person}{Kam-Fai Wong}, {and} \bibinfo{person}{Meeyoung Cha}.} \bibinfo{year}{2016}\natexlab{}.
\newblock \showarticletitle{Detecting rumors from microblogs with recurrent neural networks}.
\newblock  (\bibinfo{year}{2016}).
\newblock


\bibitem[Ma et~al\mbox{.}(2024)]%
        {ma2024simulated}
\bibfield{author}{\bibinfo{person}{Weicheng Ma}, \bibinfo{person}{Chunyuan Deng}, \bibinfo{person}{Aram Moossavi}, \bibinfo{person}{Lili Wang}, \bibinfo{person}{Soroush Vosoughi}, {and} \bibinfo{person}{Diyi Yang}.} \bibinfo{year}{2024}\natexlab{}.
\newblock \showarticletitle{Simulated misinformation susceptibility (smists): Enhancing misinformation research with large language model simulations}. In \bibinfo{booktitle}{\emph{Findings of the Association for Computational Linguistics ACL 2024}}. \bibinfo{pages}{2774--2788}.
\newblock


\bibitem[Mai et~al\mbox{.}(2023)]%
        {mai2023dynamic}
\bibfield{author}{\bibinfo{person}{Weixing Mai}, \bibinfo{person}{Zhengxuan Zhang}, \bibinfo{person}{Kuntao Li}, \bibinfo{person}{Yun Xue}, {and} \bibinfo{person}{Fenghuan Li}.} \bibinfo{year}{2023}\natexlab{}.
\newblock \showarticletitle{Dynamic graph construction framework for multimodal named entity recognition in social media}.
\newblock \bibinfo{journal}{\emph{IEEE Transactions on Computational Social Systems}} \bibinfo{volume}{11}, \bibinfo{number}{2} (\bibinfo{year}{2023}), \bibinfo{pages}{2513--2522}.
\newblock


\bibitem[Martel et~al\mbox{.}(2020)]%
        {martel2020reliance}
\bibfield{author}{\bibinfo{person}{Cameron Martel}, \bibinfo{person}{Gordon Pennycook}, {and} \bibinfo{person}{David~G Rand}.} \bibinfo{year}{2020}\natexlab{}.
\newblock \showarticletitle{Reliance on emotion promotes belief in fake news}.
\newblock \bibinfo{journal}{\emph{Cognitive research: principles and implications}}  \bibinfo{volume}{5} (\bibinfo{year}{2020}), \bibinfo{pages}{1--20}.
\newblock


\bibitem[Mridha et~al\mbox{.}(2021)]%
        {mridha2021comprehensive}
\bibfield{author}{\bibinfo{person}{Muhammad~Firoz Mridha}, \bibinfo{person}{Ashfia~Jannat Keya}, \bibinfo{person}{Md~Abdul Hamid}, \bibinfo{person}{Muhammad~Mostafa Monowar}, {and} \bibinfo{person}{Md~Saifur Rahman}.} \bibinfo{year}{2021}\natexlab{}.
\newblock \showarticletitle{A comprehensive review on fake news detection with deep learning}.
\newblock \bibinfo{journal}{\emph{IEEE access}}  \bibinfo{volume}{9} (\bibinfo{year}{2021}), \bibinfo{pages}{156151--156170}.
\newblock


\bibitem[Nan et~al\mbox{.}(2024)]%
        {nan2024let}
\bibfield{author}{\bibinfo{person}{Qiong Nan}, \bibinfo{person}{Qiang Sheng}, \bibinfo{person}{Juan Cao}, \bibinfo{person}{Beizhe Hu}, \bibinfo{person}{Danding Wang}, {and} \bibinfo{person}{Jintao Li}.} \bibinfo{year}{2024}\natexlab{}.
\newblock \showarticletitle{Let silence speak: Enhancing fake news detection with generated comments from large language models}. In \bibinfo{booktitle}{\emph{Proceedings of the 33rd ACM International Conference on Information and Knowledge Management}}. \bibinfo{pages}{1732--1742}.
\newblock


\bibitem[Perry(2023)]%
        {perry2023ai}
\bibfield{author}{\bibinfo{person}{Anat Perry}.} \bibinfo{year}{2023}\natexlab{}.
\newblock \showarticletitle{AI will never convey the essence of human empathy}.
\newblock \bibinfo{journal}{\emph{Nature Human Behaviour}} \bibinfo{volume}{7}, \bibinfo{number}{11} (\bibinfo{year}{2023}), \bibinfo{pages}{1808--1809}.
\newblock


\bibitem[Qi et~al\mbox{.}(2024)]%
        {qi2024sniffer}
\bibfield{author}{\bibinfo{person}{Peng Qi}, \bibinfo{person}{Zehong Yan}, \bibinfo{person}{Wynne Hsu}, {and} \bibinfo{person}{Mong~Li Lee}.} \bibinfo{year}{2024}\natexlab{}.
\newblock \showarticletitle{Sniffer: Multimodal large language model for explainable out-of-context misinformation detection}. In \bibinfo{booktitle}{\emph{Proceedings of the IEEE/CVF conference on computer vision and pattern recognition}}. \bibinfo{pages}{13052--13062}.
\newblock


\bibitem[Qian et~al\mbox{.}(2023)]%
        {qian2023harnessing}
\bibfield{author}{\bibinfo{person}{Yushan Qian}, \bibinfo{person}{Wei-Nan Zhang}, {and} \bibinfo{person}{Ting Liu}.} \bibinfo{year}{2023}\natexlab{}.
\newblock \showarticletitle{Harnessing the power of large language models for empathetic response generation: Empirical investigations and improvements}.
\newblock \bibinfo{journal}{\emph{arXiv preprint arXiv:2310.05140}} (\bibinfo{year}{2023}).
\newblock


\bibitem[Qu et~al\mbox{.}(2024)]%
        {qu2024qmfnd}
\bibfield{author}{\bibinfo{person}{Zhiguo Qu}, \bibinfo{person}{Yunyi Meng}, \bibinfo{person}{Ghulam Muhammad}, {and} \bibinfo{person}{Prayag Tiwari}.} \bibinfo{year}{2024}\natexlab{}.
\newblock \showarticletitle{QMFND: A quantum multimodal fusion-based fake news detection model for social media}.
\newblock \bibinfo{journal}{\emph{Information Fusion}}  \bibinfo{volume}{104} (\bibinfo{year}{2024}), \bibinfo{pages}{102172}.
\newblock


\bibitem[Rankin et~al\mbox{.}(2005)]%
        {rankin2005patterns}
\bibfield{author}{\bibinfo{person}{Katherine~P Rankin}, \bibinfo{person}{Joel~H Kramer}, {and} \bibinfo{person}{Bruce~L Miller}.} \bibinfo{year}{2005}\natexlab{}.
\newblock \showarticletitle{Patterns of cognitive and emotional empathy in frontotemporal lobar degeneration}.
\newblock \bibinfo{journal}{\emph{Cognitive and Behavioral Neurology}} \bibinfo{volume}{18}, \bibinfo{number}{1} (\bibinfo{year}{2005}), \bibinfo{pages}{28--36}.
\newblock


\bibitem[Reddy et~al\mbox{.}(2020)]%
        {reddy2020text}
\bibfield{author}{\bibinfo{person}{Harita Reddy}, \bibinfo{person}{Namratha Raj}, \bibinfo{person}{Manali Gala}, {and} \bibinfo{person}{Annappa Basava}.} \bibinfo{year}{2020}\natexlab{}.
\newblock \showarticletitle{Text-mining-based fake news detection using ensemble methods}.
\newblock \bibinfo{journal}{\emph{International journal of automation and computing}} \bibinfo{volume}{17}, \bibinfo{number}{2} (\bibinfo{year}{2020}), \bibinfo{pages}{210--221}.
\newblock


\bibitem[Ruchansky et~al\mbox{.}(2017)]%
        {ruchansky2017csi}
\bibfield{author}{\bibinfo{person}{Natali Ruchansky}, \bibinfo{person}{Sungyong Seo}, {and} \bibinfo{person}{Yan Liu}.} \bibinfo{year}{2017}\natexlab{}.
\newblock \showarticletitle{Csi: A hybrid deep model for fake news detection}. In \bibinfo{booktitle}{\emph{Proceedings of the 2017 ACM on Conference on Information and Knowledge Management}}. \bibinfo{pages}{797--806}.
\newblock


\bibitem[Shu et~al\mbox{.}(2017)]%
        {shu2017fake}
\bibfield{author}{\bibinfo{person}{Kai Shu}, \bibinfo{person}{Amy Sliva}, \bibinfo{person}{Suhang Wang}, \bibinfo{person}{Jiliang Tang}, {and} \bibinfo{person}{Huan Liu}.} \bibinfo{year}{2017}\natexlab{}.
\newblock \showarticletitle{Fake news detection on social media: A data mining perspective}.
\newblock \bibinfo{journal}{\emph{ACM SIGKDD explorations newsletter}} \bibinfo{volume}{19}, \bibinfo{number}{1} (\bibinfo{year}{2017}), \bibinfo{pages}{22--36}.
\newblock


\bibitem[Shu et~al\mbox{.}(2019)]%
        {shu2019role}
\bibfield{author}{\bibinfo{person}{Kai Shu}, \bibinfo{person}{Xinyi Zhou}, \bibinfo{person}{Suhang Wang}, \bibinfo{person}{Reza Zafarani}, {and} \bibinfo{person}{Huan Liu}.} \bibinfo{year}{2019}\natexlab{}.
\newblock \showarticletitle{The role of user profiles for fake news detection}. In \bibinfo{booktitle}{\emph{Proceedings of the 2019 IEEE/ACM international conference on advances in social networks analysis and mining}}. \bibinfo{pages}{436--439}.
\newblock


\bibitem[Singhal et~al\mbox{.}(2020)]%
        {singhal2020spotfake+}
\bibfield{author}{\bibinfo{person}{Shivangi Singhal}, \bibinfo{person}{Anubha Kabra}, \bibinfo{person}{Mohit Sharma}, \bibinfo{person}{Rajiv~Ratn Shah}, \bibinfo{person}{Tanmoy Chakraborty}, {and} \bibinfo{person}{Ponnurangam Kumaraguru}.} \bibinfo{year}{2020}\natexlab{}.
\newblock \showarticletitle{Spotfake+: A multimodal framework for fake news detection via transfer learning (student abstract)}. In \bibinfo{booktitle}{\emph{Proceedings of the AAAI conference on artificial intelligence}}, Vol.~\bibinfo{volume}{34}. \bibinfo{pages}{13915--13916}.
\newblock


\bibitem[Singhal et~al\mbox{.}(2022)]%
        {singhal2022leveraging}
\bibfield{author}{\bibinfo{person}{Shivangi Singhal}, \bibinfo{person}{Tanisha Pandey}, \bibinfo{person}{Saksham Mrig}, \bibinfo{person}{Rajiv~Ratn Shah}, {and} \bibinfo{person}{Ponnurangam Kumaraguru}.} \bibinfo{year}{2022}\natexlab{}.
\newblock \showarticletitle{Leveraging intra and inter modality relationship for multimodal fake news detection}. In \bibinfo{booktitle}{\emph{Companion Proceedings of the Web Conference 2022}}. \bibinfo{pages}{726--734}.
\newblock


\bibitem[Singhal et~al\mbox{.}(2019)]%
        {singhal2019spotfake}
\bibfield{author}{\bibinfo{person}{Shivangi Singhal}, \bibinfo{person}{Rajiv~Ratn Shah}, \bibinfo{person}{Tanmoy Chakraborty}, \bibinfo{person}{Ponnurangam Kumaraguru}, {and} \bibinfo{person}{Shin'ichi Satoh}.} \bibinfo{year}{2019}\natexlab{}.
\newblock \showarticletitle{Spotfake: A multi-modal framework for fake news detection}. In \bibinfo{booktitle}{\emph{2019 IEEE fifth international conference on multimedia big data (BigMM)}}. IEEE, \bibinfo{pages}{39--47}.
\newblock


\bibitem[Smith(2006)]%
        {smith2006cognitive}
\bibfield{author}{\bibinfo{person}{Adam Smith}.} \bibinfo{year}{2006}\natexlab{}.
\newblock \showarticletitle{Cognitive empathy and emotional empathy in human behavior and evolution}.
\newblock \bibinfo{journal}{\emph{The Psychological Record}} \bibinfo{volume}{56}, \bibinfo{number}{1} (\bibinfo{year}{2006}), \bibinfo{pages}{3--21}.
\newblock


\bibitem[Sorin et~al\mbox{.}(2024)]%
        {sorin2024large}
\bibfield{author}{\bibinfo{person}{Vera Sorin}, \bibinfo{person}{Dana Brin}, \bibinfo{person}{Yiftach Barash}, \bibinfo{person}{Eli Konen}, \bibinfo{person}{Alexander Charney}, \bibinfo{person}{Girish Nadkarni}, {and} \bibinfo{person}{Eyal Klang}.} \bibinfo{year}{2024}\natexlab{}.
\newblock \showarticletitle{Large Language Models and Empathy: Systematic Review}.
\newblock \bibinfo{journal}{\emph{Journal of Medical Internet Research}}  \bibinfo{volume}{26} (\bibinfo{year}{2024}), \bibinfo{pages}{e52597}.
\newblock


\bibitem[Steinebach et~al\mbox{.}(2019)]%
        {steinebach2019fake}
\bibfield{author}{\bibinfo{person}{Martin Steinebach}, \bibinfo{person}{Karol Gotkowski}, {and} \bibinfo{person}{Hujian Liu}.} \bibinfo{year}{2019}\natexlab{}.
\newblock \showarticletitle{Fake news detection by image montage recognition}. In \bibinfo{booktitle}{\emph{Proceedings of the 14th international conference on availability, reliability and security}}. \bibinfo{pages}{1--9}.
\newblock


\bibitem[Su et~al\mbox{.}(2020)]%
        {su2020motivations}
\bibfield{author}{\bibinfo{person}{Qi Su}, \bibinfo{person}{Mingyu Wan}, \bibinfo{person}{Xiaoqian Liu}, {and} \bibinfo{person}{Chu-Ren Huang}.} \bibinfo{year}{2020}\natexlab{}.
\newblock \showarticletitle{Motivations, methods and metrics of misinformation detection: an NLP perspective}.
\newblock \bibinfo{journal}{\emph{Natural Language Processing Research}} \bibinfo{volume}{1}, \bibinfo{number}{1} (\bibinfo{year}{2020}), \bibinfo{pages}{1--13}.
\newblock


\bibitem[Sun et~al\mbox{.}(2023)]%
        {sun2023inconsistent}
\bibfield{author}{\bibinfo{person}{Mengzhu Sun}, \bibinfo{person}{Xi Zhang}, \bibinfo{person}{Jianqiang Ma}, \bibinfo{person}{Sihong Xie}, \bibinfo{person}{Yazheng Liu}, {and} \bibinfo{person}{Philip~S Yu}.} \bibinfo{year}{2023}\natexlab{}.
\newblock \showarticletitle{Inconsistent matters: A knowledge-guided dual-consistency network for multi-modal rumor detection}.
\newblock \bibinfo{journal}{\emph{IEEE Transactions on Knowledge and Data Engineering}} \bibinfo{volume}{35}, \bibinfo{number}{12} (\bibinfo{year}{2023}), \bibinfo{pages}{12736--12749}.
\newblock


\bibitem[Tsikerdekis and Zeadally(2023)]%
        {10308424}
\bibfield{author}{\bibinfo{person}{Michail Tsikerdekis} {and} \bibinfo{person}{Sherali Zeadally}.} \bibinfo{year}{2023}\natexlab{}.
\newblock \showarticletitle{Misinformation Detection Using Deep Learning}.
\newblock \bibinfo{journal}{\emph{IT Professional}} \bibinfo{volume}{25}, \bibinfo{number}{5} (\bibinfo{year}{2023}), \bibinfo{pages}{57--63}.
\newblock
\href{https://doi.org/10.1109/MITP.2023.3314752}{doi:\nolinkurl{10.1109/MITP.2023.3314752}}


\bibitem[Wan et~al\mbox{.}(2024)]%
        {wan2024dell}
\bibfield{author}{\bibinfo{person}{Herun Wan}, \bibinfo{person}{Shangbin Feng}, \bibinfo{person}{Zhaoxuan Tan}, \bibinfo{person}{Heng Wang}, \bibinfo{person}{Yulia Tsvetkov}, {and} \bibinfo{person}{Minnan Luo}.} \bibinfo{year}{2024}\natexlab{}.
\newblock \showarticletitle{Dell: Generating reactions and explanations for llm-based misinformation detection}.
\newblock \bibinfo{journal}{\emph{arXiv preprint arXiv:2402.10426}} (\bibinfo{year}{2024}).
\newblock


\bibitem[Wang et~al\mbox{.}(2024)]%
        {wang2024misinformation}
\bibfield{author}{\bibinfo{person}{Bing Wang}, \bibinfo{person}{Ximing Li}, \bibinfo{person}{Changchun Li}, \bibinfo{person}{Bo Fu}, \bibinfo{person}{Songwen Pei}, {and} \bibinfo{person}{Shengsheng Wang}.} \bibinfo{year}{2024}\natexlab{}.
\newblock \showarticletitle{Why Misinformation is Created? Detecting them by Integrating Intent Features}. In \bibinfo{booktitle}{\emph{Proceedings of the 33rd ACM International Conference on Information and Knowledge Management}}. \bibinfo{pages}{2304--2314}.
\newblock


\bibitem[Wang et~al\mbox{.}(2022)]%
        {wang2022empathetic}
\bibfield{author}{\bibinfo{person}{Lanrui Wang}, \bibinfo{person}{Jiangnan Li}, \bibinfo{person}{Zheng Lin}, \bibinfo{person}{Fandong Meng}, \bibinfo{person}{Chenxu Yang}, \bibinfo{person}{Weiping Wang}, {and} \bibinfo{person}{Jie Zhou}.} \bibinfo{year}{2022}\natexlab{}.
\newblock \showarticletitle{Empathetic dialogue generation via sensitive emotion recognition and sensible knowledge selection}.
\newblock \bibinfo{journal}{\emph{arXiv preprint arXiv:2210.11715}} (\bibinfo{year}{2022}).
\newblock


\bibitem[Wu et~al\mbox{.}(2023)]%
        {wu2023human}
\bibfield{author}{\bibinfo{person}{Lianwei Wu}, \bibinfo{person}{Pusheng Liu}, \bibinfo{person}{Yongqiang Zhao}, \bibinfo{person}{Peng Wang}, {and} \bibinfo{person}{Yangning Zhang}.} \bibinfo{year}{2023}\natexlab{}.
\newblock \showarticletitle{Human cognition-based consistency inference networks for multi-modal fake news detection}.
\newblock \bibinfo{journal}{\emph{IEEE Transactions on Knowledge and Data Engineering}} \bibinfo{volume}{36}, \bibinfo{number}{1} (\bibinfo{year}{2023}), \bibinfo{pages}{211--225}.
\newblock


\bibitem[Wu et~al\mbox{.}(2021)]%
        {wu2021multimodal}
\bibfield{author}{\bibinfo{person}{Yang Wu}, \bibinfo{person}{Pengwei Zhan}, \bibinfo{person}{Yunjian Zhang}, \bibinfo{person}{Liming Wang}, {and} \bibinfo{person}{Zhen Xu}.} \bibinfo{year}{2021}\natexlab{}.
\newblock \showarticletitle{Multimodal fusion with co-attention networks for fake news detection}. In \bibinfo{booktitle}{\emph{Findings of the association for computational linguistics: ACL-IJCNLP 2021}}. \bibinfo{pages}{2560--2569}.
\newblock


\bibitem[Yang et~al\mbox{.}(2024)]%
        {yang2024reinforcement}
\bibfield{author}{\bibinfo{person}{Ruichao Yang}, \bibinfo{person}{Wei Gao}, \bibinfo{person}{Jing Ma}, \bibinfo{person}{Hongzhan Lin}, {and} \bibinfo{person}{Bo Wang}.} \bibinfo{year}{2024}\natexlab{}.
\newblock \showarticletitle{Reinforcement tuning for detecting stances and debunking rumors jointly with large language models}.
\newblock \bibinfo{journal}{\emph{arXiv preprint arXiv:2406.02143}} (\bibinfo{year}{2024}).
\newblock


\bibitem[Yang et~al\mbox{.}(2019)]%
        {yang2019xlnet}
\bibfield{author}{\bibinfo{person}{Zhilin Yang}, \bibinfo{person}{Zihang Dai}, \bibinfo{person}{Yiming Yang}, \bibinfo{person}{Jaime Carbonell}, \bibinfo{person}{Russ~R Salakhutdinov}, {and} \bibinfo{person}{Quoc~V Le}.} \bibinfo{year}{2019}\natexlab{}.
\newblock \showarticletitle{Xlnet: Generalized autoregressive pretraining for language understanding}.
\newblock \bibinfo{journal}{\emph{Advances in neural information processing systems}}  \bibinfo{volume}{32} (\bibinfo{year}{2019}).
\newblock


\bibitem[Yin et~al\mbox{.}({[n.\,d.]})]%
        {yindetecting}
\bibfield{author}{\bibinfo{person}{WU Yin}, \bibinfo{person}{Zhengxuan Zhang}, \bibinfo{person}{WANG Fuling}, \bibinfo{person}{Yuyu Luo}, \bibinfo{person}{Hui Xiong}, {and} \bibinfo{person}{Nan Tang}.} \bibinfo{year}{[n.\,d.]}\natexlab{}.
\newblock \showarticletitle{Detecting Out-of-Context Misinformation via Multi-Agent and Multi-Grained Retrieval}.
\newblock  (\bibinfo{year}{[n.\,d.]}).
\newblock


\bibitem[Ying et~al\mbox{.}(2023)]%
        {ying2023bootstrapping}
\bibfield{author}{\bibinfo{person}{Qichao Ying}, \bibinfo{person}{Xiaoxiao Hu}, \bibinfo{person}{Yangming Zhou}, \bibinfo{person}{Zhenxing Qian}, \bibinfo{person}{Dan Zeng}, {and} \bibinfo{person}{Shiming Ge}.} \bibinfo{year}{2023}\natexlab{}.
\newblock \showarticletitle{Bootstrapping multi-view representations for fake news detection}. In \bibinfo{booktitle}{\emph{Proceedings of the AAAI conference on Artificial Intelligence}}, Vol.~\bibinfo{volume}{37}. \bibinfo{pages}{5384--5392}.
\newblock


\bibitem[Zhang et~al\mbox{.}(2024)]%
        {zhang2024natural}
\bibfield{author}{\bibinfo{person}{Qiang Zhang}, \bibinfo{person}{Jiawei Liu}, \bibinfo{person}{Fanrui Zhang}, \bibinfo{person}{Jingyi Xie}, {and} \bibinfo{person}{Zheng-Jun Zha}.} \bibinfo{year}{2024}\natexlab{}.
\newblock \showarticletitle{Natural language-centered inference network for multi-modal fake news detection}. In \bibinfo{booktitle}{\emph{Proceedings of the Thirty-Third International Joint Conference on Artificial Intelligence, IJCAI-24}}. \bibinfo{pages}{2542--2550}.
\newblock


\bibitem[Zhou et~al\mbox{.}(2020)]%
        {zhou2020similarity}
\bibfield{author}{\bibinfo{person}{Xinyi Zhou}, \bibinfo{person}{Jindi Wu}, {and} \bibinfo{person}{Reza Zafarani}.} \bibinfo{year}{2020}\natexlab{}.
\newblock \showarticletitle{: Similarity-aware multi-modal fake news detection}. In \bibinfo{booktitle}{\emph{Pacific-Asia Conference on knowledge discovery and data mining}}. Springer, \bibinfo{pages}{354--367}.
\newblock


\bibitem[Zhou and Zafarani(2019)]%
        {zhou2019fake}
\bibfield{author}{\bibinfo{person}{Xinyi Zhou} {and} \bibinfo{person}{Reza Zafarani}.} \bibinfo{year}{2019}\natexlab{}.
\newblock \showarticletitle{Fake news detection: An interdisciplinary research}. In \bibinfo{booktitle}{\emph{Companion proceedings of the 2019 world wide web conference}}. \bibinfo{pages}{1292--1292}.
\newblock


\bibitem[Zhou and Zafarani(2020)]%
        {zhou2020survey}
\bibfield{author}{\bibinfo{person}{Xinyi Zhou} {and} \bibinfo{person}{Reza Zafarani}.} \bibinfo{year}{2020}\natexlab{}.
\newblock \showarticletitle{A survey of fake news: Fundamental theories, detection methods, and opportunities}.
\newblock \bibinfo{journal}{\emph{ACM Computing Surveys (CSUR)}} \bibinfo{volume}{53}, \bibinfo{number}{5} (\bibinfo{year}{2020}), \bibinfo{pages}{1--40}.
\newblock


\bibitem[Zhu et~al\mbox{.}(2024)]%
        {zhu2024empathizing}
\bibfield{author}{\bibinfo{person}{Jiahao Zhu}, \bibinfo{person}{Zijian Jiang}, \bibinfo{person}{Boyu Zhou}, \bibinfo{person}{Jionglong Su}, \bibinfo{person}{Jiaming Zhang}, {and} \bibinfo{person}{Zhihao Li}.} \bibinfo{year}{2024}\natexlab{}.
\newblock \showarticletitle{Empathizing Before Generation: A Double-Layered Framework for Emotional Support LLM}. In \bibinfo{booktitle}{\emph{Chinese Conference on Pattern Recognition and Computer Vision (PRCV)}}. Springer, \bibinfo{pages}{490--503}.
\newblock


\bibitem[Zollo et~al\mbox{.}(2015)]%
        {zollo2015emotional}
\bibfield{author}{\bibinfo{person}{Fabiana Zollo}, \bibinfo{person}{Petra~Kralj Novak}, \bibinfo{person}{Michela Del~Vicario}, \bibinfo{person}{Alessandro Bessi}, \bibinfo{person}{Igor Mozeti{\v{c}}}, \bibinfo{person}{Antonio Scala}, \bibinfo{person}{Guido Caldarelli}, {and} \bibinfo{person}{Walter Quattrociocchi}.} \bibinfo{year}{2015}\natexlab{}.
\newblock \showarticletitle{Emotional dynamics in the age of misinformation}.
\newblock \bibinfo{journal}{\emph{PloS one}} \bibinfo{volume}{10}, \bibinfo{number}{9} (\bibinfo{year}{2015}), \bibinfo{pages}{e0138740}.
\newblock


\bibitem[Zubiaga et~al\mbox{.}(2017)]%
        {zubiaga2017exploiting}
\bibfield{author}{\bibinfo{person}{Arkaitz Zubiaga}, \bibinfo{person}{Maria Liakata}, {and} \bibinfo{person}{Rob Procter}.} \bibinfo{year}{2017}\natexlab{}.
\newblock \showarticletitle{Exploiting context for rumour detection in social media}. In \bibinfo{booktitle}{\emph{Social Informatics: 9th International Conference, SocInfo 2017, Oxford, UK, September 13-15, 2017, Proceedings, Part I 9}}. Springer, \bibinfo{pages}{109--123}.
\newblock


\end{thebibliography}

\end{document}